\documentclass{article}

\usepackage[preprint]{aaj2026}

\usepackage[utf8]{inputenc} 
\usepackage[T1]{fontenc}    
\usepackage{hyperref}       
\usepackage{url}            
\usepackage{booktabs}       
\usepackage{amsfonts}       
\usepackage{nicefrac}       
\usepackage{microtype}      
\usepackage{xcolor}         
\usepackage{graphicx}
\usepackage{subcaption}
\usepackage{amsmath}

\title{Comparative Study of Vision-Based Metric Measurement for Large-Scale Planar Scenes}

\author{%
    ZhiXin Sun\\
  PowerChina Zhongnan Engineering Corporation Limited \\
  \texttt{sunzxjdi@gmail.com} 
}

\begin{document}

\maketitle

\begin{abstract}
Vision-based metric distance and area measurement remains a challenging problem, particularly in large-scale outdoor environments involving long-range sensing, variable camera zoom, and unstable imaging conditions. This work investigates planar metric measurement from visual observations under a real-world reservoir monitoring scenario, where the target region spans hundreds of meters and is captured by PTZ cameras. We systematically explore and compare three representative strategies, including geometry-based monocular ranging, image stitching with bird’s-eye-view transformation, and stereo-inspired ranging using two jointly calibrated monocular cameras. For monocular ranging, planar localization models are derived based on camera imaging geometry, and the sensitivity of measurement accuracy to camera pitch angle is analyzed. To support large-area measurement, an image mosaicking framework integrating coarse bird’s-eye-view transformation, image retrieval, feature matching, homography estimation, and bundle adjustment optimization is further investigated. In addition, a stereo-based ranging scheme is developed to achieve long-range metric estimation without dedicated stereo hardware. Experimental results demonstrate distinct trade-offs among these approaches. Monocular ranging achieves meter-level accuracy when the pitch angle is sufficiently large, while the stereo-based strategy achieves decimeter-level accuracy and is considerably less sensitive to camera pitch angle compared with monocular ranging. Image stitching exhibits strong robustness for small-scale scenes but becomes increasingly unstable and computationally demanding as scene size and image quantity grow. These findings provide practical insights for selecting appropriate vision-based measurement strategies in large-scale planar monitoring applications. Code is available at \url{https://github.com/sunzx97/Vision_Based_Distance_Measurement.git}
\end{abstract}

\section{Introduction}
Vision-based distance and area measurement is a long-standing yet still challenging problem, particularly in large-scale outdoor scenarios involving long-range sensing and zoom-capable cameras. This work investigates planar point distance measurement using visual information. Under the planar-scene assumption, both monocular and binocular configurations become applicable for metric distance estimation, with monocular approaches particularly benefiting from the geometric constraint. Although recent advances in computer vision have led to extensive research on monocular depth estimation~~\cite{lin2025depth, wang2025vggt, raj2025neoarcade, bochkovskiy2025depth}, traditional methods based on camera imaging geometry often provide higher metric accuracy in constrained scenarios~\cite{yamaguti1997method, dandil2019computer, s22030962}. Our application scenario is an outdoor reservoir with dimensions of 220m×300m, where two cameras are installed beside the reservoir. The distance from the cameras to the water surface is approximately 8–12 m. The reservoir freezes in winter, and our goal is to estimate the ice-covered area based on camera observations. The key task is to compute the distance from specified points on the ice surface to the camera. The cameras are variable-zoom PTZ devices. We explore several approaches, including monocular ranging, image stitching, and binocular stereo measurement, which are detailed in the following sections.

\section{Methods}
\begin{figure}[htbp]
    \centering
    \includegraphics[width=\linewidth]{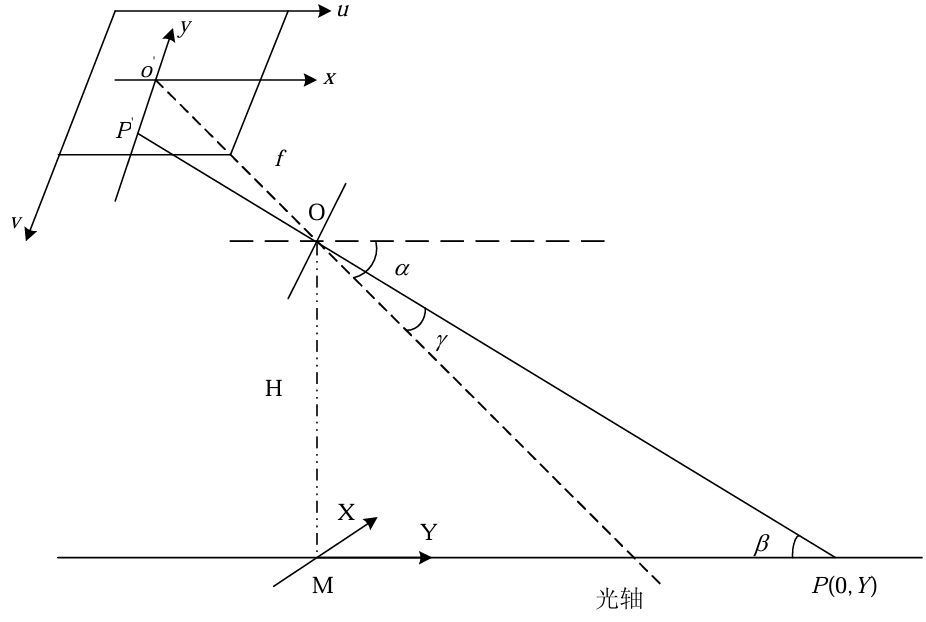}
    \caption{Vertical Geometric Localization Model of a Planar Point.}
    \label{camera_model1}
\end{figure}

Given the camera intrinsic parameters, camera installation pitch angle, and camera height, the distance from the camera to a point on the ground plane can be estimated. To simplify the geometric modeling of monocular distance measurement, following~\cite{wang2012monocular}, we first establish the vertical localization model of a planar point, as illustrated in Fig.~\ref{camera_model1}.

Assume that the camera focal length is $f$, the camera installation height is $H$, and the pitch angle is $\alpha$. The projection point $P'$ has image-plane coordinates $(0,y)$ and corresponding pixel coordinates $(u_0,v)$. The image center is denoted as $(u_0,v_0)$. According to the geometric relationship, the longitudinal distance $Y$ from the camera to the planar point can be expressed as

\begin{equation}
Y=\frac{H}{\tan\beta}
=\frac{H}{\tan(\alpha-\gamma)}
=\frac{H}{\tan\left(\alpha-\arctan\left(\frac{y}{f}\right)\right)},
\label{eq:Y1}
\end{equation}

where $\gamma$ is the angle between the optical axis and the projection ray. Considering the relationship between image-plane and pixel coordinates,

\begin{equation}
y=(v-v_0)d_y,
\label{eq:y_pixel}
\end{equation}

where $d_y$ denotes the physical distance represented by one pixel along the vertical axis. Substituting Eq.~\ref{eq:y_pixel} into Eq.~\ref{eq:Y1}, we obtain

\begin{equation}
Y
=
\frac{H}
{\tan\left(
\alpha-
\arctan\left(
\frac{(v-v_0)d_y}{f}
\right)
\right)}
=
\frac{H}
{\tan\left(
\alpha-
\arctan\left(
\frac{v-v_0}{f_y}
\right)
\right)},
\label{eq:Y}
\end{equation}

where

\begin{equation}
f_y=\frac{f}{d_y}.
\end{equation}

\begin{figure}[htbp]
    \centering
    \includegraphics[width=\linewidth]{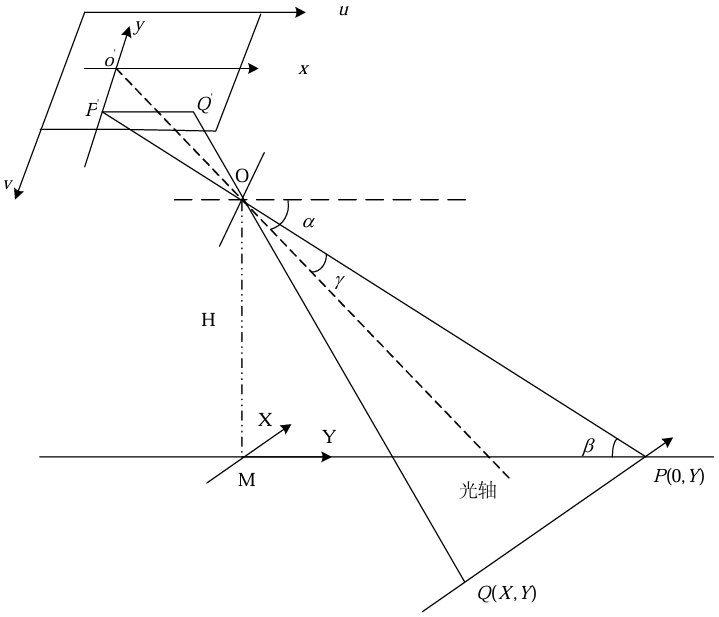}
    \caption{Horizontal Geometric Localization Model of a Planar Point.}
    \label{camera_model2}
\end{figure}

Next, we extend the formulation to the horizontal localization model of planar points, as shown in Fig.~\ref{camera_model2}. Assume that the planar point $Q$ has world coordinates $(X,Y)$ and lies on the same horizontal line as point $P$. Its projection point $Q'$ has image-plane coordinates $(x,y)$ and corresponding pixel coordinates $(u,v)$.

From the similarity relationship $\triangle OP'Q' \sim \triangle OPQ$, we have

\begin{equation}
\frac{OP}{OP'}
=
\frac{PQ}{P'Q'}.
\label{eq:similar}
\end{equation}

In right triangle $OO'P'$,

\begin{equation}
OP'
=
\sqrt{O'O^2+O'P'^2}
=
\sqrt{f^2+y^2}
=
\sqrt{
f^2+(v-v_0)^2d_y^2
}.
\label{eq:OPprime}
\end{equation}

Similarly, in right triangle $OMP$,

\begin{equation}
OP
=
\frac{OM}{\sin\beta}
=
\frac{H}
{\sin\left(
\alpha-
\arctan\left(
\frac{y}{f}
\right)
\right)}
=
\frac{H}
{\sin\left(
\alpha-
\arctan\left(
\frac{v-v_0}{f_y}
\right)
\right)}.
\label{eq:OP}
\end{equation}

The horizontal image coordinate satisfies

\begin{equation}
x=(u-u_0)d_x,
\end{equation}

where $d_x$ denotes the physical spacing represented by one pixel along the horizontal axis. Therefore, according to Eq.~\ref{eq:similar}, the horizontal coordinate of the planar point can be derived as

\begin{equation}
X
=
PQ
=
P'Q'
\cdot
\frac{OP}{OP'}
=
(u-u_0)d_x
\cdot
\frac{
\displaystyle
\frac{H}
{\sin\left(
\alpha-
\arctan\left(
\frac{v-v_0}{f_y}
\right)
\right)}
}
{
\sqrt{
f^2+(v-v_0)^2d_y^2
}
}.
\label{eq:X1}
\end{equation}

By introducing the horizontal focal length in pixel units,

\begin{equation}
f_x=\frac{f}{d_x},
\end{equation}

Eq.~\ref{eq:X1} can be further simplified as

\begin{equation}
X
=
\frac{
(u-u_0)H
}
{
\sqrt{
f_x^2+(v-v_0)^2
}
\,
\sin
\left(
\alpha-
\arctan
\left(
\frac{v-v_0}{f_y}
\right)
\right)
}.
\label{eq:X}
\end{equation}

\begin{figure}[htbp]
    \centering
    \includegraphics[width=\linewidth]{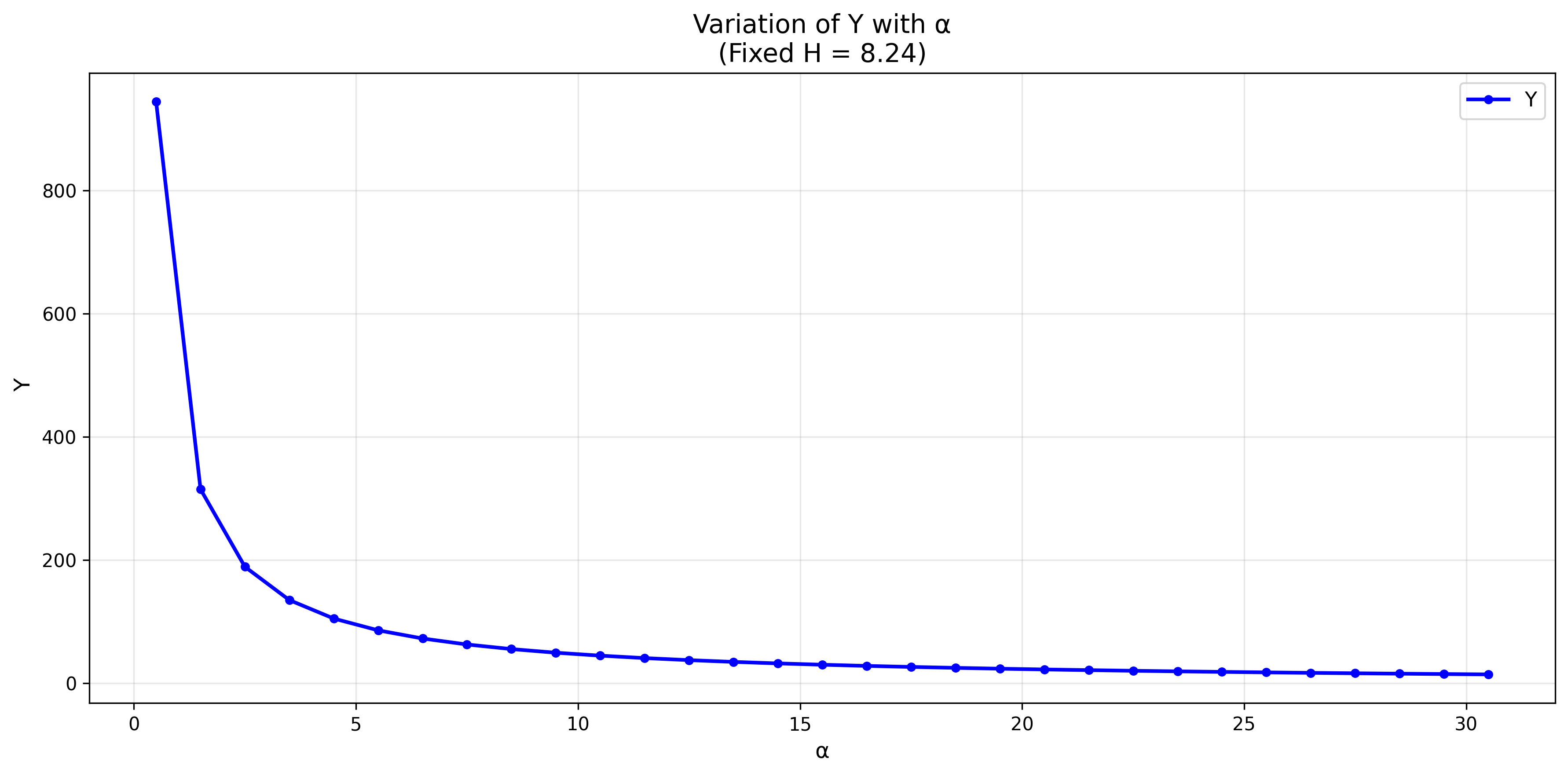}
    \caption{Sensitivity of estimated distance $Y$ to pitch angle $\alpha$ (Fixed H = 8.24)}
    \label{triangle_side_variation}
\end{figure}

To improve the accuracy of outdoor measurements, it is necessary to calibrate several known points to correct for the pitch angle deviation caused by camera installation and the misalignment between the optical center and the pixel coordinate center. Furthermore, the measurement error increases drastically when the pitch angle is extremely small. In our measurement example, the camera height is $H=8.24\,\text{m}$. When the pitch angle of a point in the vertical direction is $2.679^\circ$, the calculated $Y$ is $171.61\,\text{m}$; however, when the pitch angle decreases to $1.679^\circ$, the calculated $Y$ surges to $281.11\,\text{m}$. Factors such as camera installation errors, wind-induced vibrations, and the optical center offset inevitably introduce deviations in the pitch angle. Figure~\ref{triangle_side_variation} intuitively illustrates the variation of $Y$ as the pitch angle $\alpha$ ranges from $0.5^\circ$ to $30^\circ$ with $H=8.24\,\text{m}$. Consequently, this method becomes unsuitable when $\alpha$ is too small.

\begin{figure}[htbp]
    \centering
    \includegraphics[width=\linewidth]{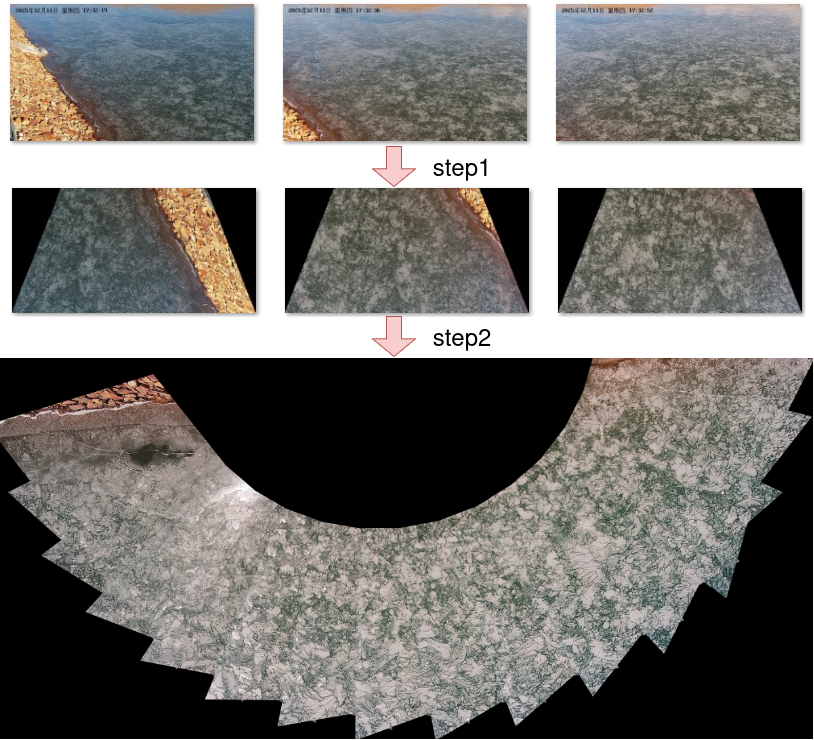}
    \caption{Overview of the proposed image stitching framework. Each image is first transformed into a coarse bird's-eye view using camera geometry, followed by neighboring image retrieval, LightGlue-based feature matching, homography estimation, and global optimization via bundle adjustment (BA).}
    \label{stich}
\end{figure}

An image mosaicking strategy was further investigated. In principle, if the entire planar scene can be stitched into a single image and more than four control points with known coordinates are calibrated around the plane, perspective distortion can be completely eliminated. Under this framework, each camera pose can be transformed into a bird's-eye view through the homography matrix estimated for the corresponding image. Figure~\ref{stich} illustrates our image stitching pipeline.

Each image is first transformed into a coarse bird's-eye view. The transformation is termed ``coarse'' because it relies on camera intrinsic and extrinsic parameters obtained directly from the SDK, which inevitably introduce calibration inaccuracies, especially under variable zoom settings where accurate focal-length-specific calibration is difficult.

After obtaining these coarse bird's-eye views, we employ NetVLAD~\cite{arandjelovic2016netvlad, sarlin2019coarse} to identify potentially overlapping image sets, followed by feature matching using LightGlue~\cite{lindenberger2023lightglue}, from which pairwise homography matrices are estimated. Finally, bundle adjustment (BA) is applied for global optimization. Notably, the initial bird's-eye-view transformation step can be omitted. We included this step mainly because, at the early stage of our work, BA-based optimization was not considered, and our initial approach relied on pairwise image stitching with progressively expanding planar coverage.

Experimental results indicate that image stitching is feasible for small-scale regions. However, in large-scale scenarios requiring thousands of images, geometric errors accumulate progressively due to lens distortion and imperfect homography estimation. In addition, bundle adjustment may converge to local optima, reducing global consistency and potentially causing reconstruction failure. Furthermore, in outdoor environments, cameras are continuously affected by wind-induced vibrations. Even when the camera is commanded to return to the same PTZ position, discrepancies often remain between the images used for stitching and those captured during testing.

Compared with pure image stitching, 3D reconstruction frameworks offer the advantage of jointly optimizing camera poses and lens distortion parameters. Nevertheless, they remain sensitive to camera motion and environmental instability.

\begin{figure}[htbp]
    \centering
    \includegraphics[width=\linewidth]{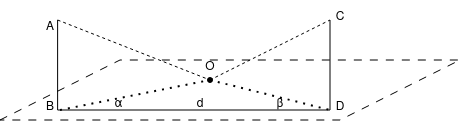}
    \caption{Geometry of the proposed stereo-based distance measurement method using two monocular cameras. The baseline distance $d$ between the cameras is measured by RTK, while the target distance is estimated from the camera yaw angles and the law of sines under a bird's-eye-view geometry.}
    \label{two}
\end{figure}

A stereo-based distance measurement strategy was finally investigated. Commercial stereo ranging cameras are often limited in measurable distance due to their relatively short baseline. However, based on the geometric principles of stereo vision, we can achieve a stereo ranging effect by jointly calibrating two monocular cameras. As illustrated in Figure~\ref{two}, cameras $AB$ and $CD$ denote two monocular cameras separated by a baseline distance $d$, which can be accurately measured using RTK. Let $\alpha$ and $\beta$ represent the yaw angles of the two cameras. Under the bird's-eye-view perspective, the sides $OB$ and $OD$ of triangle $OBD$ can be derived using the law of sines.

Define
\begin{equation}
\gamma = \pi - \alpha - \beta .
\end{equation}
According to the law of sines,
\begin{equation}
\frac{OD}{\sin\alpha}
=
\frac{OB}{\sin\beta}
=
\frac{d}{\sin\gamma},
\end{equation}
thus yielding
\begin{equation}
OD
=
\frac{d\sin\alpha}{\sin\gamma},
\qquad
OB
=
\frac{d\sin\beta}{\sin\gamma}.
\end{equation}
The yaw angles $\alpha$ and $\beta$ are obtained by combining the camera yaw angles provided by the SDK with the pixel-level yaw angles within the image. Assuming camera focal lengths $f_x$ and $f_y$, and principal point coordinates $(u_0,v_0)$, the pixel-level yaw angle corresponding to an image point $(u,v)$ is computed as
\begin{equation}
\text{yaw}
=
\arctan
\left(
\frac{u-u_0}{f_x}
\right).
\end{equation}

In our practical experiments, this approach achieved the highest ranging accuracy among the explored methods. Moreover, the method is broadly applicable to scenarios requiring distance estimation between two points under a bird's-eye-view perspective. For example, it can be used to assess whether suspended loads from tower cranes maintain a safe distance from personnel on the ground.

\section{Experiments}
According to our experimental results, monocular distance measurement can achieve meter-level accuracy when the camera pitch angle is no less than $30^\circ$. To maintain the measurement error within 5\%, stereo-based ranging can further improve the accuracy to the decimeter level. 
Image stitching demonstrates strong robustness and can achieve accurate mosaicking when the number of images does not exceed approximately 40. However, as the number of images increases, the computational cost grows substantially, and the stitching process may become unstable or even fail.

\section{Conclusion}
This work investigated vision-based metric measurement for large-scale planar scenes under a real-world outdoor reservoir monitoring scenario. Three representative approaches were explored and comparatively analyzed, including geometry-based monocular ranging, image stitching with bird’s-eye-view transformation, and stereo-based ranging using two calibrated monocular cameras.
Experimental results reveal clear trade-offs among the investigated methods. Monocular ranging provides a simple and computationally efficient solution, but its accuracy is highly sensitive to camera pitch angle and calibration errors, particularly in long-range scenarios with small viewing angles. Image stitching enables large-area planar mapping and perspective correction, yet suffers from accumulated geometric errors, high computational cost, and reduced robustness when the number of images becomes large or camera instability is present. In contrast, the stereo-based ranging strategy demonstrates the most favorable performance in our experiments, achieving decimeter-level accuracy while maintaining practical applicability for long-distance measurement tasks.
Although the current study focuses on planar measurement under outdoor conditions, the proposed analysis provides broader insights into vision-based metric sensing using monocular and binocular configurations. Future work will investigate jointly optimized geometric calibration and 3D reconstruction frameworks, as well as more robust methods for handling camera motion, lens distortion, and large-scale scene reconstruction under continuously varying PTZ conditions.
\bibliography{biblio.bib}

\end{document}